\documentclass{article}
\usepackage{preprint,times}

\usepackage{enumitem}
\usepackage{amsmath,amssymb,booktabs,graphicx}

\usepackage[hidelinks]{hyperref}
\usepackage{xcolor}
\usepackage{tcolorbox}
\usepackage{tikz}
\usepackage{array}
\usepackage{siunitx}

\usepackage{colortbl}
\usepackage{multirow}
\usepackage{wrapfig}
\usepackage{needspace}
\usepackage[below]{placeins}
\newcolumntype{C}{>{\centering\arraybackslash}p{1.35cm}}

\definecolor{OliveGreen}{cmyk}{0.64,0,0.95,0.40}
\definecolor{BurntOrange}{rgb}{0.8, 0.33, 0.0}

\definecolor{best}{RGB}{210, 235, 255}   %
\definecolor{second}{RGB}{255, 240, 210} %

\newcommand{\best}[1]{\cellcolor{best}\textbf{#1}}

\title{Understanding Confabulation and Rethinking Reconstruction in Activation Explanations}

\author{%
  \begin{minipage}[t]{\dimexpr\textwidth-2\tabcolsep\relax}
    \centering\normalfont\normalsize
    {\bfseries
      Gert Lek\textsuperscript{1}\hspace{1.2em}%
      Zixuan Xia\textsuperscript{1,2}\hspace{1.2em}%
      Pin-Yu Chen\textsuperscript{3}\hspace{1.2em}%
      Lydia Y. Chen\textsuperscript{1,4}}\\[5pt]
    \begin{tabular}{@{}l@{\hspace{2em}}l@{}}
      \textsuperscript{1} Universit\'e de Neuch\^atel &
      \textsuperscript{2} Universit\"at Bern \\
      \textsuperscript{3} IBM Research &
      \textsuperscript{4} Delft University of Technology
    \end{tabular}\\[3pt]
    \texttt{gert.lek@unine.ch}
  \end{minipage}%
}

\begin{document}
\maketitle

\suppressfloats[t]

\begin{abstract}
Natural Language Autoencoders (NLAs) produce unsupervised text explanations of a model's activations: a verbalizer describes an activation and a reconstructor learns to recover it from this text.
Under the established point-reconstruction NLA training recipe, explanations become more useful for predicting model behavior while also increasingly introducing unsupported details and exhibiting writing defects.
To assess these changes separately, we introduce a standardized evaluation framework for unstructured NLA explanations, measuring information recoverable from explanations, contextual support for their claims, and writing quality.
To address confabulation and writing defects, we move beyond predicting a single activation: explanations can distinguish distributions of possible activations even when their means and optimal point-reconstruction rewards are identical.
We introduce Flow-NLA, which models the distribution of activations compatible with an explanation and trains the verbalizer using a diffusion likelihood bound.
Across Qwen, Gemma, and Apertus, this richer signal retains the utility gains of point reconstruction while curbing the growth of confabulation and writing defects, opening up a direction for improving activation-derived training to encourage more informative, supported, and readable explanations.
Code and evaluation prompts will be made publicly available upon acceptance.
\end{abstract}

\section{Introduction}
\label{sec:outline-introduction}

Natural-language activation explanations aim to make model representations accessible beyond observable outputs, as explored by LatentQA and Activation Oracles~\citep{pan2026latentqa,karvonen2025activationoracles}.
Natural Language Autoencoders (NLAs)~\citep{frasertaliente2026nla}
learn such explanations through reconstruction: a verbalizer describes an activation in natural language, and a reconstructor predicts the activation from that text.
After supervised initialization, reconstruction quality provides reinforcement-learning feedback without requiring labels for each generated explanation.

Because removing true claims hurts reconstruction more than removing false ones, the original NLA study anticipated that training would reduce confabulation, but found that it remained substantial and did not decrease~\citep{frasertaliente2026nla}.
Rerunning this recipe on three open models, we find that final explanations are similarly confabulated and that confabulation rises as reconstruction improves.
Recent work has further examined initialization sensitivity and claim-level reliability~\citep{zhang2026nlarobustness,dingeto2026decodability}.
These findings raise a fundamental question: \emph{what properties of an explanation does reconstruction training actually reward?}
An explanation may convey useful information while introducing unsupported claims or expressing that information poorly.
We therefore separate these properties, track how they evolve during training, and ask whether richer reconstruction feedback can improve explanation quality beyond point reconstruction.

Our contributions are:
\begin{enumerate}[
    label=\textbf{\arabic*.},
    wide,
    labelindent=0pt,
    itemsep=3pt,
    topsep=4pt
]

\item \textbf{Evaluating explanation quality.}
We measure instance-specific predictive information by asking an
independent reader model to make predictions from the explanation,
comparing matched explanations with mismatched explanations.
Separately, we evaluate contextual confabulation, claims unsupported or
contradicted by the source context using claim-based
evaluation~\citep{min2023factscore,song2024veriscore}, and identify writing
defects in clarity, coherence, and economy
(Section~\ref{sec:outline-evaluation}).
This framework separates what information an explanation makes recoverable
from whether its claims are context-supported and clearly expressed.

\item \textbf{Characterizing point-reconstruction training.}
Prior NLA work primarily trains explanations through point reconstruction.
Across the models we study, this training improves predictive utility
while also increasing contextual confabulation and writing defects
(Section~\ref{sec:training-trajectories}).
This behavior follows naturally from the training signal:
reconstruction rewards text for helping recover the target activation, but
does not directly require the claims carrying that information to be
context-supported or clearly expressed.
Thus, useful but unsupported cues can still be rewarded, motivating a
closer examination of what information the reconstruction objective itself
captures.

\item \textbf{Beyond point reconstruction with Flow-NLA.}
We show that point reconstruction has an information blind spot:
under NLA's normalized squared-error objective, explanation refinements that
change the conditional activation distribution while preserving its mean
cannot improve the optimal reconstruction reward.
Flow-NLA instead provides explanation-conditioned denoising feedback across
noise levels, allowing the training signal to depend on information beyond
the conditional mean (Section~\ref{sec:outline-method}).
We relate this objective to the diffusion variational
bound~\citep{ho2020denoising,kingma2021variational} and analyze its
sensitivity to richer conditional information. Across three models, Flow-NLA retains substantial utility gains while limiting contextual confabulation growth and producing fewer writing defects than the baseline NLAs (Section~\ref{sec:outline-results}).
\end{enumerate}

\section{Related Work}
\label{sec:outline-related}
\textbf{Learning Activation Explanations.}
\label{sec:related-learning}
LatentQA and Activation Oracles train language models to answer questions about activations using supervised activation--response pairs~\citep{pan2026latentqa,karvonen2025activationoracles}, with Activation Oracles broadening this supervision through classification and context-prediction tasks.
More recent approaches use reconstruction to learn open-ended activation descriptions beyond predefined questions.
Cycle-Consistent Activation Oracles~\citep{chalnev2026cycleconsistent}
warm-start the verbalizer on LatentQA-style supervision and
surrounding-token prediction before optimizing an
activation$\rightarrow$text$\rightarrow$activation cycle with RL, while NLAs~\citep{frasertaliente2026nla} warm-start both the activation verbalizer and reconstructor using teacher-generated descriptions before jointly optimizing them for activation reconstruction.
These approaches make reconstruction a learning signal for open-ended
activation explanations, but the feedback remains centered on recovering
individual target activations.
We instead study whether richer, distributional reconstruction feedback
can reward information that point reconstruction misses, using
conditional denoising to examine its effects on explanation utility and
confabulation.

\textbf{Confabulation and Explanation Evaluation.}
\label{sec:related-confabulation}
An NLA initialization study finds that claim plausibility declines during training for normally initialized NLAs and that different initializations yield similar reconstruction accuracy despite differences in explanation plausibility~\citep{zhang2026nlarobustness}.
Verbalizations can also reflect the verbalizer's own parametric
knowledge~\citep{li2026privileged}. For utility, CHIVE evaluates whether
interpretability tools, including NLAs, improve counterfactual behavioral
prediction beyond the original prefix~\citep{karvonen2026chive}.
The NLA study evaluates both information recovery and confabulation~\citep{frasertaliente2026nla}.
We extend this evaluation effort with a standardized framework for unstructured NLA explanations, enabling comparisons across models and training procedures.
Our framework evaluates predictions from explanations with the source
withheld and separately checks source support using claim-based
evaluation adapted from FActScore and VeriScore~\citep{min2023factscore,song2024veriscore}.

\section{Confabulation or Explanation? Understanding Confabulation in Reconstruction Training}
\label{sec:outline-finding}
We start by examining how explanations change during training as an NLA learns to reconstruct activations more accurately.
To achieve this, we propose an evaluation framework that scores utility, confabulation, and writing quality.

First, we describe the established reconstruction training procedure. Let $h\in\mathbb R^d$ denote a residual-stream
activation of the target language model at a selected layer and token position, normalized to unit
length. The target language model is frozen. We extract activations from fixed residual-stream locations, with layers and token positions specified for each target model and evaluation task in Appendix~\ref{sec:activation-extraction}.
The verbalizer
$\pi_\phi(z\mid h)$ generates an explanation $z$ from the activation.
Following the released NLA~\citep{frasertaliente2026nla}, a separate reconstructor predicts a unit
activation $\widehat h_\theta(z)$, discarding activation magnitude. The verbalizer's reward is
$
    r_\theta(h,z)=-\left\|h-\widehat h_\theta(z)\right\|_2^2.
$
Training an NLA consists of two phases. After supervised initialization on activation--summary pairs, online training
repeatedly generates groups of explanations, scores them with the current reconstructor, and updates the reconstructor and the verbalizer on the same batch~\citep{frasertaliente2026nla}.
The reconstructor is updated directly on the generated activation--explanation pairs.
The verbalizer uses a group-relative policy update with a KL penalty toward its
supervised reference~\citep{shao2024deepseekmath}. Thus, an explanation
is favored when the reconstructor can use it to reduce the reconstruction MSE.

\subsection{An Evaluation Framework for Explanation Quality}
\label{sec:outline-evaluation}

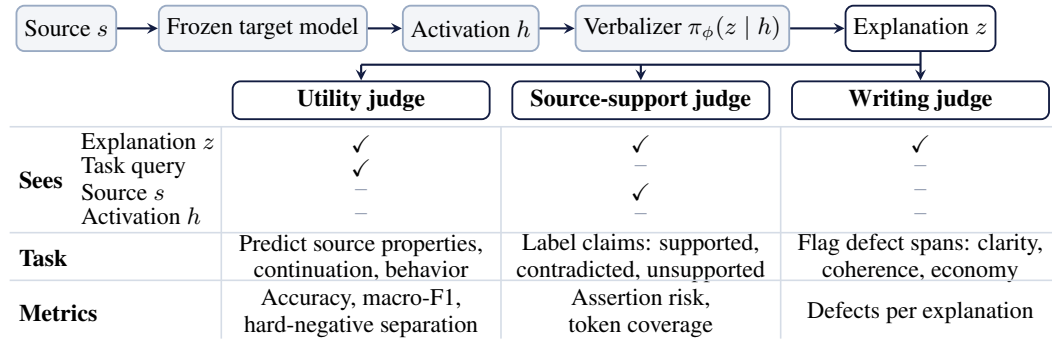
\begin{figure*}[b]
    \centering
    \definecolor{evNavy}{HTML}{121C41}
\definecolor{evFill}{HTML}{F3F5F8}
\definecolor{evLine}{HTML}{99AAC3}
\definecolor{evRule}{HTML}{D5DCE6}
\definecolor{evMuted}{HTML}{9AA3B5}
\begin{tikzpicture}[
    x=1cm,y=1cm,>=stealth,
    every node/.style={font=\small,text=black,align=center},
    flowbox/.style={rectangle,rounded corners=3pt,line width=0.7pt,
        draw=evLine,fill=evFill,minimum height=0.6cm,inner sep=3pt},
    judge/.style={rectangle,rounded corners=3pt,line width=0.7pt,
        draw=evNavy,fill=white,minimum width=3.4cm,minimum height=0.5cm,inner sep=2pt,
        font=\small\bfseries},
    rowlabel/.style={anchor=west,font=\footnotesize},
    group/.style={anchor=west,font=\footnotesize\bfseries},
    cell/.style={font=\footnotesize},
    signal/.style={->,draw=evNavy,line width=0.8pt},
    rule/.style={draw=evRule,line width=0.6pt}
]
\def\cA{4.65}\def\cB{8.35}\def\cC{12.05}
\def\yes{\checkmark}
\def\no{\textcolor{evMuted}{--}}

\node[flowbox] (source) at (0.75,0) {Source $s$};
\node[flowbox] (target) at (3.35,0) {Frozen target model};
\node[flowbox] (act) at (6.1,0) {Activation $h$};
\node[flowbox] (verb) at (8.9,0) {Verbalizer $\pi_\phi(z\mid h)$};
\node[flowbox,draw=evNavy,fill=white] (expl) at (\cC,0) {Explanation $z$};
\draw[signal] (source) -- (target);
\draw[signal] (target) -- (act);
\draw[signal] (act) -- (verb);
\draw[signal] (verb) -- (expl);

\node[judge] (utility) at (\cA,-0.95) {Utility judge};
\node[judge] (support) at (\cB,-0.95) {Source-support judge};
\node[judge] (writing) at (\cC,-0.95) {Writing judge};
\draw[draw=evNavy,line width=0.8pt] (\cC,-0.48) -- (\cA,-0.48);
\draw[signal] (expl.south) -- (writing.north);
\draw[signal] (\cA,-0.48) -- (utility.north);
\draw[signal] (\cB,-0.48) -- (support.north);

\draw[rule] (0,-1.3) -- (13.9,-1.3);
\node[group] at (0,-2.0) {Sees};
\foreach \y/\label/\a/\b/\c in {
    -1.52/Explanation $z$/\yes/\yes/\yes,
    -1.84/Task query/\yes/\no/\no,
    -2.16/Source $s$/\no/\yes/\no,
    -2.48/Activation $h$/\no/\no/\no} {
    \node[rowlabel] at (0.82,\y) {\label};
    \node at (\cA,\y) {\a};
    \node at (\cB,\y) {\b};
    \node at (\cC,\y) {\c};
}

\draw[rule] (0,-2.7) -- (13.9,-2.7);
\node[group] at (0,-3.03) {Task};
\node[cell] at (\cA,-3.03) {Predict source properties,\\continuation, behavior};
\node[cell] at (\cB,-3.03) {Label claims: supported,\\contradicted, unsupported};
\node[cell] at (\cC,-3.03) {Flag defect spans: clarity,\\coherence, economy};
\draw[rule] (0,-3.33) -- (13.9,-3.33);
\node[group] at (0,-3.76) {Metrics};
\node[cell] at (\cA,-3.76) {Accuracy, macro-F1,\\hard-negative separation};
\node[cell] at (\cB,-3.76) {Assertion risk,\\token coverage};
\node[cell] at (\cC,-3.76) {Defects per explanation};
\foreach \x in {2.8,6.5,10.2}
    \draw[rule] (\x,-1.3) -- (\x,-4.1);
\end{tikzpicture}
    \caption{
        \textbf{Three judges of explanation quality.}
        Each judge sees only what it requires: The utility judge
        answers the task query from the explanation; the source-support
        judge checks the explanation's claims with the source; and the
        writing judge assesses only the explanation.
    }
    \label{fig:evaluation-overview}
\end{figure*}

We evaluate explanations along three complementary dimensions
(Figure~\ref{fig:evaluation-overview}): \textbf{utility}, measuring what
task-relevant information a blind judge can recover from the explanation;
\textbf{source support}, measuring whether its claims are supported by the
source; and \textbf{writing quality}, measuring how clearly that information
is expressed. 
These dimensions are scored by a separate judge LLM, whose context is restricted to what each dimension requires. The utility judge answers a task query using only the explanation, the source-support judge observes both the source and explanation, and the writing judge observes only the explanation. 
Appendix~\ref{sec:main3-evaluation-details} provides further information.

\paragraph{Utility.}

Useful explanations let readers answer diverse questions about the activation. Therefore, we evaluate utility over a large range of tasks (Appendix Table~\ref{tab:main3-utility}).
First, we identify properties of the source text through 12 explanation classification tasks spanning semantic (topic, sentiment, subjectivity, entailment), lexical (entity and exact-name presence), and grammatical (tense, number, language, pronoun form) categories (accuracy). 
Second, we extract information on the model's continuation by classifying the true 32-token continuation among ten candidates (accuracy).
Last, we predict the model's own behavior by predicting whether it answers, refuses, or asks for clarification from the explanation (macro-F1). 
Explanations that perform well across these tasks carry identifiable information from the activation. The utility judge receives only the explanation and a question with candidate answers (the task query), and must choose an answer. These are judged with DeepSeek V4.1 Flash~\citep{deepseekai2026v41flash} as the judge model (Appendix~\ref{sec:main3-evaluation-details}).

To additionally test whether utility comes from instance-specific information rather than coarse cues such as the topic, we compare the matched explanation against a hard negative:  the explanation of a closely related source that differs in the relevant answer. 
For example, ``Alice lent Bob a book'' and ``Bob lent Alice a book'' share nearly all
surface cues but imply opposite answers to ``Who lent the book?'' 
We define the hard-negative separation as the log-probability ratio
$\log\bigl[p_{\mathrm J}(y\mid z_{\mathrm{matched}})/p_{\mathrm J}(y\mid z_{\mathrm{hard}})\bigr]$
that the utility judge assigns to the correct answer $y$ given the matched versus the hard-negative explanation, with answer probabilities read from the same DeepSeek judge; it is the primary utility measure presented here.
Higher is better: in the example, an explanation that only says a book was lent scores zero, while one that says who lent it to whom scores positive; positive values thus indicate instance-specific information beyond the topic and surface cues shared with the hard negative.
Definitions and aggregation details are given in
Appendix~\ref{sec:main3-evaluation-details}, and
Appendix~\ref{sec:hard-negative-details} decomposes the separation into
matched predictive gain and mismatch penalty.

\paragraph{Confabulation.} 
Using a source-support judge that receives the source text before the activation extraction and the explanation, explanations are split 
into distinct atomic claims \citep{min2023factscore,song2024veriscore}. The claims are labeled as supported (stated in or entailed by the source), contradicted (incompatible with the source) or unsupported. 
Supported and contradicted verdicts are required to quote the exact source evidence. 
This yields two metrics: Assertion Risk, which is the proportion of assertions that are unsupported or contradicted, and Confabulated Token Coverage, which is the proportion of explanation tokens occupied by these claims.
Both metrics assess the contextual source support rather than the utility of describing the model's internal
computation.

\paragraph{Writing quality.}
The writing quality judge sees only the explanation, without the source text.
Here, the judge flags distinct defects in clarity (unclear wording, malformed or unfinished sentences), coherence
(statements that conflict within the explanation), and economy (avoidable
repetition or filler), quoting the exact explanation span for each.  We count
distinct defects and report changes from supervised initialization. Writing quality is evaluated separately from
source support because an explanation may be unsupported yet well written, or supported yet poorly expressed.

For example, the Qwen NLA explanation below fails the source-support and writing quality judges for different reasons: the source-support judge flags invented names as contradicted by the source, and the writing judge flags incompatible accounts of the person's name as a coherence defect.

\begin{tcolorbox}[
colback=gray!3,
colframe=gray!40,
boxrule=0.5pt,
arc=2pt,
top=2pt,
bottom=2pt
]
\textbf{Source}\\
Jonn Hart is a stage name his real name is Di'Jonn Grizzell

\smallskip
\textbf{Explanation excerpt}\\
The final clause ``his actual full name is Kristopher Ellis...
his real last name is Scott Gellell; he's actual full name is Tracy
Ellzell'' requires a period or continuation
\end{tcolorbox}

\begin{figure}[!b]
    \centering
    \includegraphics[
        width=\linewidth,
        trim={11bp 11bp 12bp 0bp},
        clip=true
    ]{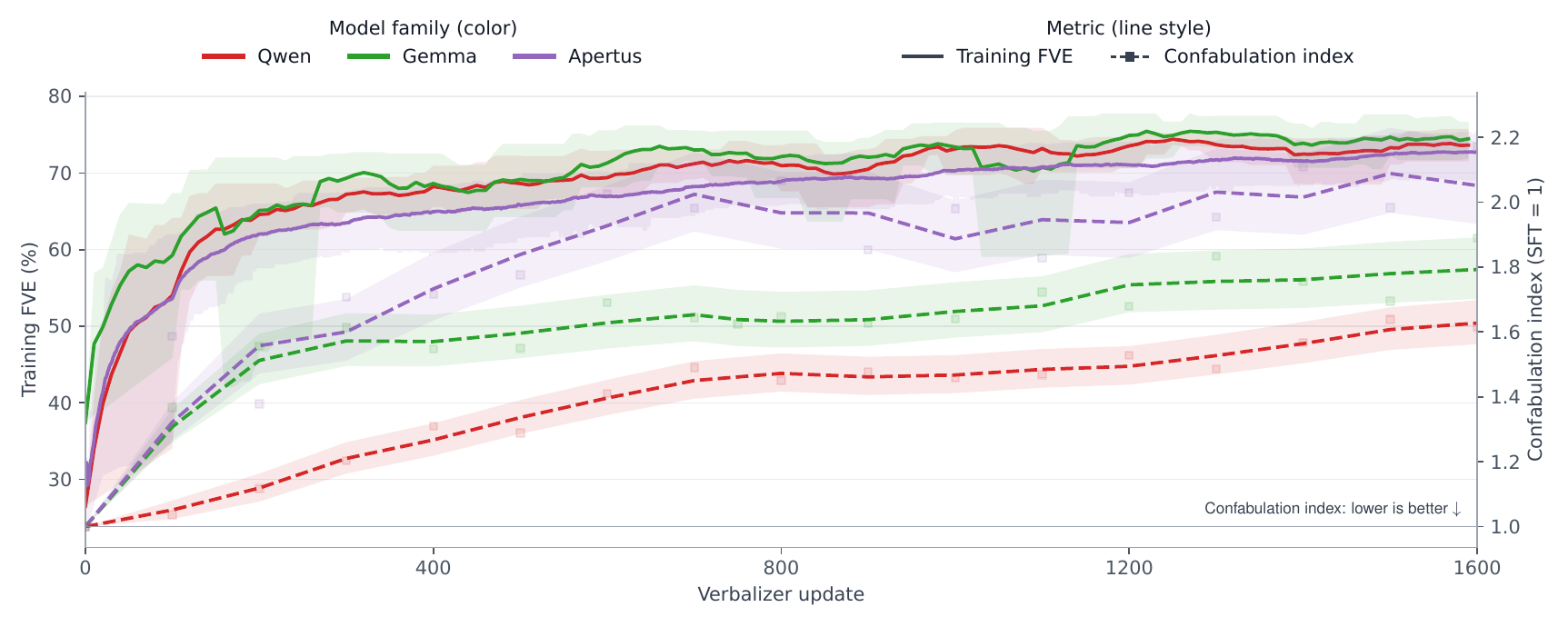}
    \caption{\textbf{Better reconstruction accompanies more confabulation and writing defects.}
    As the FVE reconstruction rises (solid line), so does the confabulation index (dashed line) across models.
    The index $I(t)=\tfrac12[R(t)/R(0)+D(t)/D(0)]$ averages assertion risk $R$ and
    writing defects per explanation $D$ relative to supervised initialization
    ($I=1$). Shading shows 95\% bootstrap intervals.}
    \label{fig:combined-defects-risk}
\end{figure}

\subsection{The Confabulation Cost of Better Reconstruction}
\label{sec:training-trajectories}
We retrain NLAs using the established setup from \citet{frasertaliente2026nla} for Qwen2.5-7B-Instruct~\citep{qwen2025qwen25}, Gemma-3-12B~\citep{gemmateam2025gemma3} and Apertus-v1.5-8B~\citep{apertus2026}.
Reconstruction is measured during training using the Fraction of Variance Explained, $\mathrm{FVE}=1-\mathbb{E}[\|h-\widehat h_\theta(z)\|_2^2]/V_h$, where $V_h$ is the variance baseline for unit activations; utility, confabulation, and writing quality are evaluated on held-out examples.

As reconstruction improves during training, assertion risk and writing defects increase (Figure~\ref{fig:combined-defects-risk}; separate panels in Appendix~\ref{sec:individual-evaluation-panels}): the correlation between FVE and the confabulation index is considerable, with 0.78 for Qwen, 0.96 for Gemma, and 0.90 for Apertus.

\begin{wrapfigure}{r}{0.5\linewidth}
    \centering
    \vspace{-5pt} %
    \includegraphics[width=\linewidth]{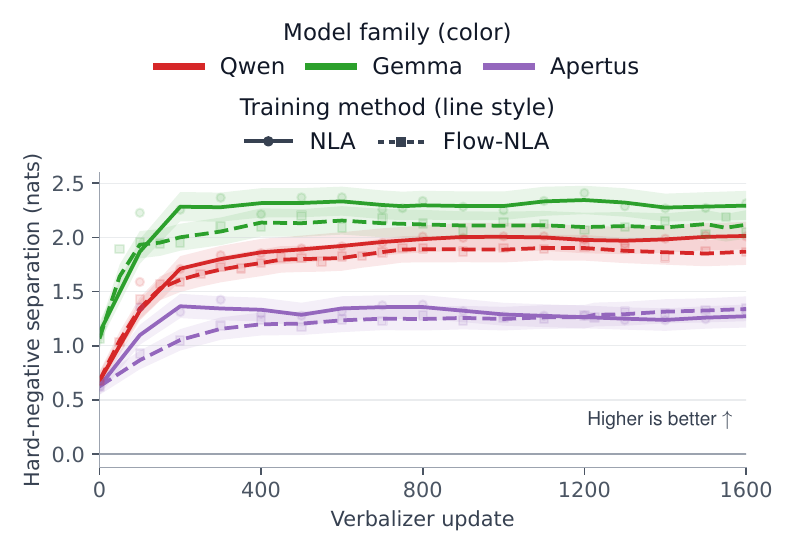}
    \caption{\textbf{Better reconstruction accompanies more relevant
    information.} Hard-negative separation across the 12 classification datasets
    for point-reconstruction NLA (solid) and Flow-NLA (dashed); higher is better.}
    \label{fig:hard-negative-separation}
    \vspace{-18pt}
\end{wrapfigure}

At the same time, hard-negative separation rises from roughly 0.6--1.1 nats at supervised initialization to 1.2--2.4 nats within the first few hundred updates (Figure~\ref{fig:hard-negative-separation}), so explanations increasingly carry instance-specific information rather than only coarse cues (Section~\ref{sec:outline-evaluation}).
Better reconstruction thus carries both greater recoverable information and a deterioration in how reliably and clearly it is expressed, although whether this coupling is strictly necessary remains open.
\Needspace{24\baselineskip}
\paragraph{Released NLA baselines.}
The same coexistence of useful information and unsupported content is
visible in the four released Anthropic NLAs~\citep{frasertaliente2026nla}
(Table~\ref{tab:baseline-comparison}). Classification accuracy ranges
from 79.4\% to 85.7\% and suffix-selection accuracy from 96.6\% to 99.0\%,
yet assertion risk exceeds 75\% across all four models, and confabulated
content accounts for more than half of explanation tokens in the released NLAs.

Classification and suffix-selection accuracies approximately match those reported by
\citet{frasertaliente2026nla} for their Claude NLAs
(about 78--86\% and 93--98\%), although their evaluation code is
unreleased.  Our model reruns reach a training FVE of about 0.75
(Figure~\ref{fig:combined-defects-risk}), close to the 0.75--0.80
reported for the released checkpoints. Our point-reconstruction reruns
on Qwen, Gemma, and Apertus show the same pattern: 77.5\% to 84.2\%
classification and 96.0\% to 97.6\% suffix-selection accuracy,
alongside 77.8\% to 83.7\% assertion risk and 41.5\% to 68.3\%
confabulated token coverage. Thus, both released checkpoints and our
reruns produce useful but highly confabulated explanations. We next
ask whether changing the activation-derived reward can preserve
useful learning while weakening the coupling between confabulation
and reconstruction error.
\par
\ifnum\value{WF@wrappedlines}>1
    \vspace{\dimexpr\value{WF@wrappedlines}\baselineskip-\baselineskip\relax}
\fi
\csname WFclear\endcsname
\begin{table}[t]
    \caption{\textbf{Utility and confabulation in released NLAs and our reruns.}
    All scores are percentages except writing defects, which are counts per
    explanation.}
    \label{tab:baseline-comparison}
    \centering
    \setlength{\tabcolsep}{2pt}
    \begin{tabular}{@{}l@{\hspace{9pt}}l
        S[table-format=2.1] S[table-format=2.1] S[table-format=2.1]
        S[table-format=2.1] S[table-format=2.1] S[table-format=1.2]@{}}
        \toprule
        & & \multicolumn{3}{c}{Utility $\uparrow$}
        & \multicolumn{2}{c}{Confabulation $\downarrow$}
        & {Writing $\downarrow$} \\
        \cmidrule(lr){3-5}\cmidrule(lr){6-7}\cmidrule(l){8-8}
        & {Target model} & {Class.} & {Suffix} & {Behavior}
        & {Risk} & {Coverage} & {Defects} \\
        \midrule
        \multirow{4}{*}{\footnotesize\itshape\shortstack[l]{Released\\Anthropic NLAs}} & Qwen2.5-7B     & 79.4 & 98.0 & 82.7 & 92.6 & 74.8 & 2.45 \\
        & Gemma-3-12B    & 84.7 & 96.6 & 83.4 & 83.1 & 67.5 & 1.78 \\
        & Gemma-3-27B    & 85.7 & 98.2 & 84.1 & 76.3 & 55.4 & 1.92 \\
        & Llama-3.3-70B  & 82.7 & 99.0 & 65.1 & 76.4 & 56.9 & 1.78 \\
        \midrule
        \multirow{3}{*}{\footnotesize\itshape\shortstack[l]{Point-reconstruction\\reruns}} & Qwen2.5-7B     & 79.7 & 97.6 & 74.9 & 83.7 & 68.3 & 1.63 \\
        & Gemma-3-12B    & 84.2 & 96.2 & 82.4 & 77.8 & 44.2 & 1.38 \\
        & Apertus-v1.5-8B & 77.5 & 96.0 & 73.6 & 81.3 & 41.5 & 1.50 \\
        \bottomrule
    \end{tabular}
    \vspace{-6pt}
\end{table}

\section{Beyond Point Reconstruction with Flow-NLA}
\label{sec:outline-method}
What information should reconstruction feedback reward?
We first show that informative distinctions among activations can leave optimal point-reconstruction loss unchanged.
Next, we develop Flow-NLA to reward such distinctions through conditional denoising and evaluate how this richer feedback changes explanation utility, confabulation, and writing quality.

\subsection{Beyond Point Reconstruction}
\label{sec:point-reconstruction}

Point reconstruction can fail to reward informative distinctions even
with an optimal reconstructor. For a fixed verbalizer, the joint
distribution $p(h)\pi_\phi(z\mid h)$ induces $p_\phi(h\mid z)$.
Following NLA~\citep{frasertaliente2026nla}, let both the target $H$ and
prediction $u$ have unit norm. Writing
$\mu_\phi(z)=\mathbb E_\phi[H\mid Z=z]$, we have
\begin{equation}
\min_{\|u\|_2=1}
\mathbb E_\phi[\|H-u\|_2^2\mid Z=z]
=2-2\|\mu_\phi(z)\|_2.
\label{eq:main3-point-optimum}
\end{equation}
The optimum is attained by $u=\mu_\phi(z)/\|\mu_\phi(z)\|_2$
when $\mu_\phi(z)\ne0$, and by any unit-norm prediction otherwise.
The optimal expected point loss depends on $p_\phi(h\mid z)$ only
through its conditional mean. Because NLA rewards the verbalizer based on reconstruction quality, more informative explanations that leave the conditional mean unchanged do not necessarily receive higher reward.

\begin{figure}[t]
\centering
\includegraphics[width=\linewidth,trim=8 8 7 8,clip]{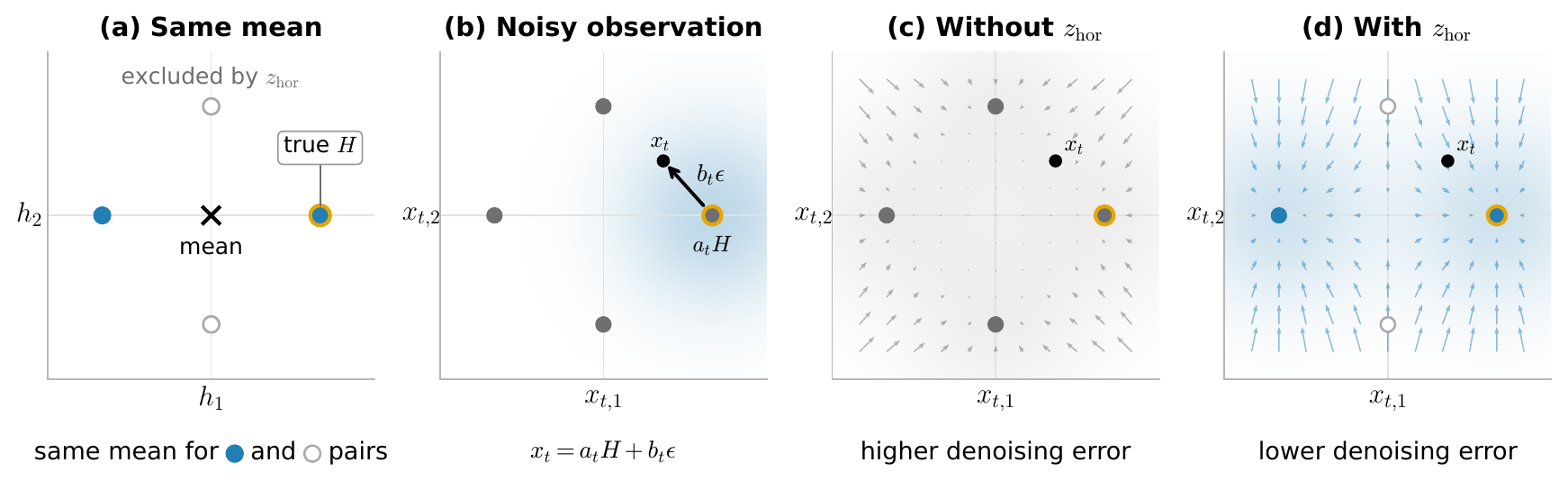}
\caption{
\textbf{Denoising can reward information beyond the conditional mean.}
$H$ is uniform over four unit directions; we follow the realized activation $H=(1,0)$ (gold outline).
(a) The horizontal explanation $z_{\mathrm{hor}}$ keeps the horizontal pair
(blue) and excludes the vertical pair (open circles). Both pairs have mean
zero ($\times$), so the optimal expected point loss is the same with or
without $z_{\mathrm{hor}}$.
(b) Denoising also observes the noisy activation $x_t=a_tH+b_t\epsilon$
(black arrow: noise $b_t\epsilon$), with candidates scaled by $a_t$.
(c) Without the explanation, $x_t$ is about equally close to the true
and upward candidates, so the optimal estimate hedges between them.
(d) With $z_{\mathrm{hor}}$, the upward candidate is excluded and the
estimate moves to the true direction, lowering the denoising error.
In (c)--(d), arrows show the displacement
$a_t\mathbb E_\phi[H\mid x_t,z]-x_t$ from each noisy observation to its
optimal signal estimate (without $z$ in (c)); shading shows the density
of $x_t$.
}
\label{fig:equal-mean-reconstruction}
\end{figure}

Figure~\ref{fig:equal-mean-reconstruction} illustrates this limitation
with four equally likely unit directions. The horizontal explanation
$z_{\mathrm{hor}}$ rules out the vertical directions but
leaves two opposing candidates. Their mean is zero, so this partial
information leaves the optimal expected point loss at $2$, as
without the explanation.

Denoising additionally observes $x_t=a_tH+b_t\epsilon$, where
$\epsilon\sim\mathcal N(0,I)$ is independent of $(H,Z)$ and
$a_t^2+b_t^2=1$. At each noise level, the optimal unconstrained
squared-error prediction is $\mathbb E_\phi[H\mid x_t,z]$.
The explanation can therefore exclude candidates that remain plausible
from $x_t$ alone. For the illustrated observation, denoising assigns
lower error with the correct horizontal explanation than without it
(Figure~\ref{fig:equal-mean-reconstruction}c--d).
Information that leaves the conditional mean unchanged can thus still be rewarded by denoising.

\subsection{Flow-NLA}
\label{sec:flow-nla}
\begin{figure}[!b]
    \centering
    \includegraphics[width=\linewidth]{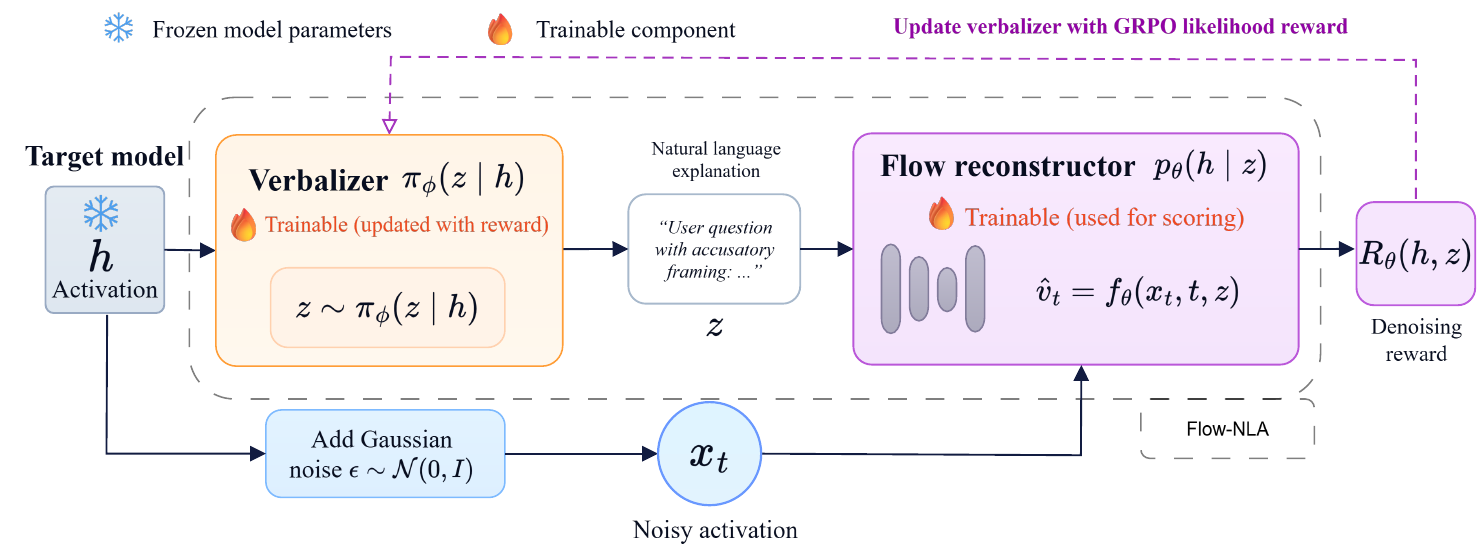}
    \caption{\textbf{Learning activation explanations with Flow-NLA.}
    An activation $h$ from a frozen target model is passed to a verbalizer,
    which generates an explanation $z$, and corrupted with Gaussian noise
    to form $x_t$. The flow reconstructor learns to denoise $x_t$
    conditioned on $z$, modeling the distribution of activations compatible
    with the explanation. Denoising errors across noise levels define a
    reward derived from a conditional diffusion likelihood bound.
    This reward trains the verbalizer to produce explanations that make
    the observed activation more predictable under the conditional model.}
    \label{fig:method-overview}
\end{figure}
Flow-NLA builds on this argument by replacing point reconstruction with conditional denoising.
The reconstructor learns to denoise activations using the text explanation, and the verbalizer is rewarded for explanations that help it capture the conditional distribution of activations they describe.
This results in an unsupervised training signal sensitive to more than the conditional mean. 

Rewarding input reconstruction likelihood under a co-trained decoder is an established way to encourage informative outputs~\citep{barber2003im,miao2016language,eysenbach2019diayn,rita2022emergent}, typically using categorical or Gaussian models. Flow-NLA instead rewards input likelihood under a co-trained diffusion model conditioned on the verbalizer's output. This conditional distribution shifts as the verbalizer learns, complicating reward stabilization.
This unsupervised reward favors activation prediction over source-context support, so it may still reward unsupported wording. Source-text supervision could address this, but would bias explanations away from information extractable from activations alone.
Whether this objective changes the balance between utility and contextual
confabulation is an empirical question
(Section~\ref{sec:outline-results}).

\paragraph{Conditional reconstruction.}
Figure~\ref{fig:method-overview} shows the Flow-NLA training procedure.
Let $x$ be the activation after a fixed coordinate standardization,
and let $X$ denote the corresponding random variable. We form a noisy
activation $x_t=a_tx+b_t\epsilon$ and velocity target
$v_t=a_t\epsilon-b_tx$, where $\epsilon\sim\mathcal N(0,I_d)$ and
$a_t^2+b_t^2=1$ define a variance-preserving path, and $d$ is the activation dimension.
The reconstructor $f_\theta(x_t,t,z)$ combines a text reader with a
conditional velocity field and is trained with the velocity-prediction
objective~\citep{salimans2022progressive}:
\begin{equation}
    \mathcal L_{\mathrm{rec}}(\theta;\phi)
    =\mathbb E_{h,z,t,\epsilon}
       \left[d^{-1}\|f_\theta(x_t,t,z)-v_t\|_2^2\right],
    \qquad z\sim\pi_\phi(\cdot\mid h).
    \label{eq:flow-loss}
\end{equation}

Across noise levels, the model learns how the explanation helps distinguish activation patterns compatible with a partially observed activation.
These denoising predictions capture conditional distributional structure~\citep{saharia2021sr3}, since their population optimum at an interior noise level identifies the full explanation-conditioned distribution for a fixed verbalizer (Appendix~\ref{sec:statistical-foundations}).
During training, this distribution,
$p_\phi(h\mid z)\propto p(h)\,\pi_\phi(z\mid h)$, changes whenever the
verbalizer learns, so the reconstructor tracks a
moving target, much like a distributional
critic~\citep{bellemare2017distributional}, except that here the modeled
distribution itself, not its mean, defines the reward
(Appendix~\ref{sec:reward-stability}).

\paragraph{Rewarding distributional information.}
We seek to reward explanations under which the observed activation has high conditional likelihood.
The point reward already does this, but with a restrictive model: for unit
vectors, $r_\theta(h,z)=2\,\widehat h_\theta(z)^{\top}h-2$ is, up to scale and
a constant, the log-likelihood of a distribution on the unit sphere
determined entirely by its mean direction $\widehat h_\theta(z)$.
Such a log-likelihood lower-bounds the information
that explanations carry about activations~\citep{barber2003im}. The bound
is loose whenever the model cannot represent the true conditional
distribution, as with the two opposing candidates in
Figure~\ref{fig:equal-mean-reconstruction}. Flow-NLA keeps this objective
but replaces this model with the flow reconstructor, which
can represent several modes.
The diffusion likelihood bound~\citep{ho2020denoising,kingma2021variational,kingma2023understanding}
provides a tractable training signal: it relates this likelihood to how
well the explanation helps the reconstructor denoise the activation
across noise levels.
The bound takes the form
$\log p_\theta(h\mid z)\ge R_\theta(h,z)-c(h)$
(Appendix~\ref{sec:likelihood-bound}),
where $R_\theta(h,z)$ is the explanation-dependent denoising term and
$c(h)$ is independent of the explanation. We therefore use $R_\theta$
as the verbalizer's reward, so that candidate explanations of the same
activation are compared through their likelihood bounds.
Flow matching trains the reconstructor to predict the velocity $v_t$, whereas the
bound is expressed in noise-prediction error. The definitions of $x_t$ and $v_t$ imply
$b_tx_t+a_tv_t=(a_t^2+b_t^2)\epsilon=\epsilon$. Substituting the predicted
velocity therefore gives
$\widehat\epsilon_\theta(x_t,t,z)=b_tx_t+a_tf_\theta(x_t,t,z)$,
without training a second predictor.

\begin{figure}[t]
    \centering
    \includegraphics[width=\linewidth,trim=24 16 2 0,clip]{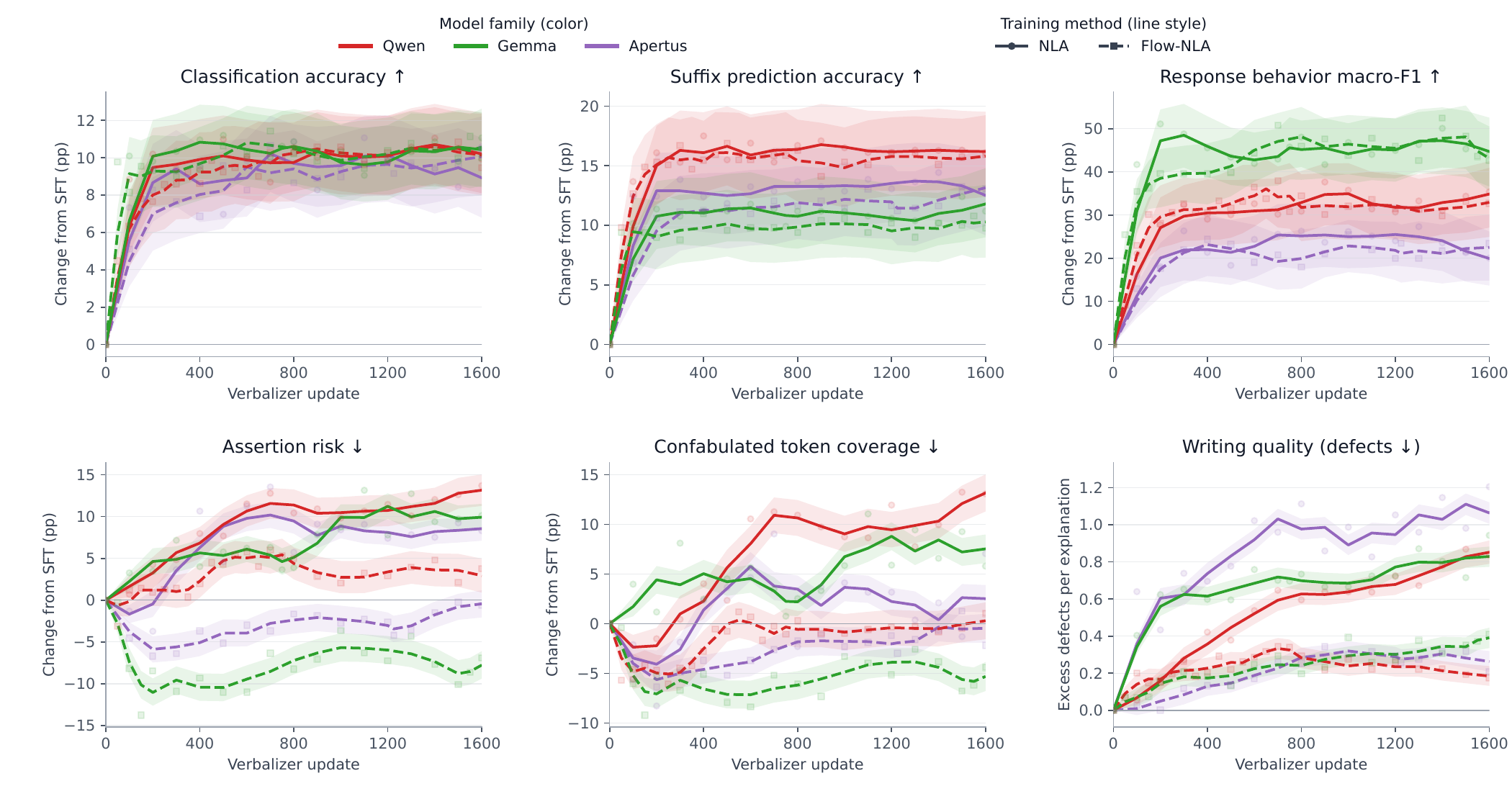}
    \caption{Changes in utility, confabulation, and writing defects from
    supervised initialization for point-reconstruction NLA (solid) and
    Flow-NLA (dashed). Markers show raw measurements, curves show three-checkpoint means,
    and shading shows pointwise 95\% bootstrap intervals.}
    \label{fig:richer-feedback-results}
\end{figure}

We index noise levels by log signal-to-noise ratio,
$\lambda=\log(a_t^2/b_t^2)$: low values correspond to mostly noise and high
values to mostly signal. In this coordinate, the likelihood-bound term
integrates noise-prediction error with constant weight $1/2$
(Appendix~\ref{sec:likelihood-bound}). We use
$\lambda_{\min}=-8$ and $\lambda_{\max}=8$, a finite band spanning both
regimes. The reward is the negative integrated noise-prediction error:
\begin{equation}
    R_\theta(h,z)
    =-\frac12\int_{\lambda_{\min}}^{\lambda_{\max}}
        \mathbb E_\epsilon
        \left\|\epsilon-\widehat\epsilon_\theta(x_\lambda,\lambda,z)
        \right\|_2^2\,\mathrm d\lambda.
    \label{eq:distributional-reward}
\end{equation}
For optimal denoisers, each integrand is the clean-activation denoising error of Figure~\ref{fig:equal-mean-reconstruction} scaled by $e^{\lambda}$, since $\epsilon-\widehat\epsilon^{*}=e^{\lambda/2}(\mathbb E[X\mid x_\lambda,z]-x)$.
The reward therefore credits an explanation for every noise level at which it makes the observed activation easier to recover, including when the explanation leaves the conditional mean unchanged; in the example of Figure~\ref{fig:equal-mean-reconstruction}, $R_\theta$ favors $z_{\mathrm{hor}}$ over an uninformative explanation by approximately $\log 2$ nats, while the point reward $r_\theta$ is unchanged. This information-theoretic interpretation motivates constant weighting in log-SNR: it matches the likelihood-bound term, whereas arbitrary reweightings do not (Appendix~\ref{sec:information-theory}). The trained reconstructor and numerical integration approximate this ideal quantity~\citep{kingma2021variational}.

\paragraph{Training the verbalizer.}
We follow the standard NLA training recipe described in
Section~\ref{sec:outline-finding}, retaining the verbalizer's supervised initialization and group-relative updates with a KL penalty~\citep{frasertaliente2026nla}. The reconstructor's supervised initialization differs, as it is trained for velocity prediction rather than point reconstruction.
We replace the point-reconstruction reward with
$R_\theta(h,z)$, estimated using sampled Gaussian noise and numerical
integration over log-SNR $[-8,8]$, and fit the online reconstructor to the
generated activation--explanation pairs using the denoising objective, in
the same order as the standard recipe.
Because the distribution behind this reward moves with the verbalizer,
several choices mattered for training (Appendix~\ref{sec:reward-stability}).
Integrating over noise levels avoids uninformative extremes: nearly clean inputs reveal the activation regardless of the explanation, while nearly pure noise leaves only the conditional mean (Appendix~\ref{sec:statistical-foundations}). To assess whether the reconstructor captures more distributional information as the verbalizer learns, we compare sample and real activation covariance within and outside the activations' 64 leading principal components (Appendix~\ref{sec:distributional-fit}). This diagnostic ruled out our initial energy-score reward~\citep{gneiting2007proper}, which compares the observed activation with reverse-generated samples: their spread outside the leading subspace decreased as mean reconstruction improved. The likelihood-bound reward requires no reverse-generated samples, only forward noising of the observed activation, and its reconstructor's spread outside the leading subspace converges to that of the activations.
\begin{figure}[t]
    \centering
    \includegraphics[width=\linewidth]{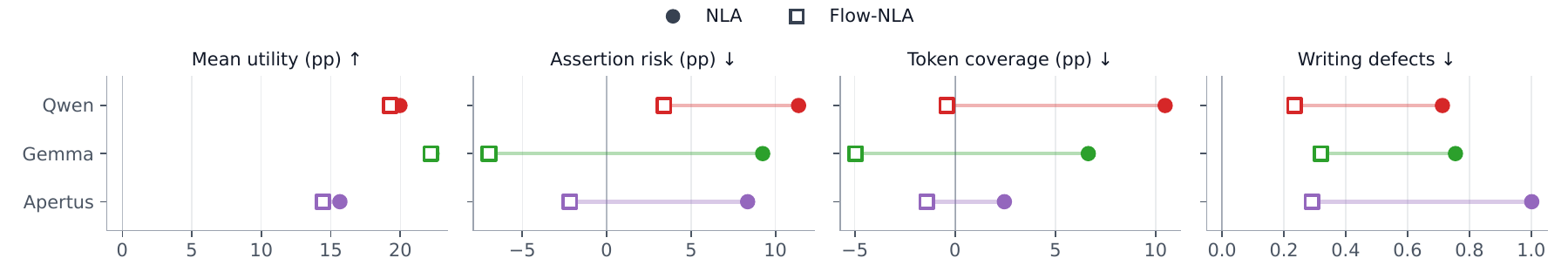}
    \caption{Changes from supervised initialization in mean utility,
    assertion risk, token coverage, and writing defects, averaged over the
    second half of the training range shared by both methods. Filled circles
    denote NLA; open squares denote Flow-NLA; lines show the gap between
    methods.}
    \label{fig:flow-vs-nla-dumbbell}
    \vspace{-5pt}
\end{figure}
\subsection{Utility and Confabulation under Richer Feedback}
\label{sec:outline-results}
We evaluate whether Flow-NLA changes the balance between recoverable
information and explanation defects using the evaluation framework from
Section~\ref{sec:outline-evaluation}.
We compare Flow-NLA with point-reconstruction NLA on
Qwen2.5-7B-Instruct, Gemma-3-12B, and Apertus-v1.5-8B, examining whether
the effect of richer feedback is consistent across LLMs.
For each model, we compare both methods' training trajectories in
utility, contextual confabulation, and writing quality
(Figure~\ref{fig:richer-feedback-results}).

On all three target models, the two methods follow similar utility
trajectories, while confabulation and writing defects grow significantly less
under Flow-NLA. Figure~\ref{fig:flow-vs-nla-dumbbell} summarizes this late
in training: Flow-NLA retains the mean utility gain (averaged over classification, suffix, and behavior), while its assertion risk and token coverage stay near or below
supervised initialization rather than rising by up to 11 points, and its
writing defects grow by 0.23--0.32 rather than 0.71--1.00 per explanation.
Flow-NLA also has the lowest assertion risk, token coverage, and writing defects of all compared NLAs, including the
released ones (Table~\ref{tab:flow-model-comparison}, Appendix~\ref{sec:absolute-model-comparison}).

Across target models, substantial utility gains can thus be retained with less growth in confabulation and defects, establishing richer reconstruction feedback as a promising direction for NLA training.

\FloatBarrier
\section{Conclusions, Limitations, and Future Work}
Current NLAs grow less reliable as they learn: during point-reconstruction
training, explanations become more useful but also more confabulated and less
clearly written, and the released NLAs show the same pattern. Our evaluation
framework exposes this by scoring utility, source support, and writing quality
separately. This deterioration points to a limitation of the current objective:
point reconstruction rewards whatever text helps recover the activation,
while remaining blind to much of the information an explanation carries.
Flow-NLA replaces it with a conditional denoising reward that credits this
richer information. Across three target models, Flow-NLA keeps its
utility gains while sharply curbing the growth of confabulation and writing defects during training.

Our study is limited to the models, tasks, and training procedures considered, relies on automated judgments, and includes provider and inference differences that limit causal attribution to the reconstruction objective alone (Appendix~\ref{sec:main3-evaluation-details}). Our method and analysis also assume optimal denoisers and a fixed verbalizer, and richer reconstruction feedback does not guarantee source support or mechanistic faithfulness. Future work should better characterize how reconstruction rewards induce confabulation and design training objectives that encourage informative, supported explanations.

\newpage

\bibliographystyle{preprint_refs}
\bibliography{references}

\appendix
\section{Absolute Model Comparison}
\label{sec:absolute-model-comparison}

\begin{table}[!ht]
    \caption{\textbf{Utility and confabulation across NLA training methods.}
    We compare released NLAs, our point-reconstruction reruns, and Flow-NLA
    within the same target model. All scores are percentages except writing
    defects, which are counts per explanation. Best results are highlighted within each target model.
    Checkpoint selection is described in
    Appendix~\ref{sec:baseline-comparison-details}.}
    \label{tab:flow-model-comparison}
    \begin{center}
    \setlength{\tabcolsep}{2.5pt}
    \begin{tabular}{@{}l@{\hspace{9pt}}l
        S[table-format=2.1] S[table-format=2.1] S[table-format=2.1]
        S[table-format=2.1] S[table-format=2.1] S[table-format=1.2]@{}}
        \toprule
        & & \multicolumn{3}{c}{Utility $\uparrow$}
        & \multicolumn{2}{c}{Confabulation $\downarrow$}
        & {Writing $\downarrow$} \\
        \cmidrule(lr){3-5}\cmidrule(lr){6-7}\cmidrule(l){8-8}
        {Target model} & {Method}
        & {Class.} & {Suffix} & {Behavior}
        & {Risk} & {Coverage} & {Defects} \\
        \midrule

        \multirow{3}{*}{Qwen2.5-7B}
        & Released
        & 79.4 & \best{98.0} & \best{82.7}
        & 92.6 & 74.8 & 2.45 \\

        & Point recon.
        & 79.7 & 97.6 & 74.9
        & 83.7 & 68.3 & 1.63 \\

        & Flow-NLA
        & \best{80.3} & 96.6 & 72.9
        & \best{71.9} & \best{54.7} & \best{0.99} \\

        \midrule

        \multirow{3}{*}{Gemma-3-12B}
        & Released
        & \best{84.7} & \best{96.6} & 83.4
        & 83.1 & 67.5 & 1.78 \\

        & Point recon.
        & 84.2 & 96.2 & 82.4
        & 77.8 & 44.2 & 1.38 \\

        & Flow-NLA
        & 83.8 & 96.2 & \best{84.8}
        & \best{60.8} & \best{33.8} & \best{0.89} \\

        \midrule

        \multirow{2}{*}{Apertus-v1.5-8B}
        & Point recon.
        & 77.5 & \best{96.0} & \best{73.6}
        & 81.3 & 41.5 & 1.50 \\

        & Flow-NLA
        & \best{78.4} & 95.4 & 69.6
        & \best{72.8} & \best{38.9} & \best{0.75} \\

        \bottomrule
    \end{tabular}
    \end{center}
\end{table}

\section{Evaluation Details}
\label{sec:main3-evaluation-details}

\subsection{Activation extraction}
\label{sec:activation-extraction}
Table~\ref{tab:activation-extraction} lists the extraction site for each
target model. As in the original NLA code~\citep{frasertaliente2026nla}, our reruns and Flow-NLA share the extraction site of their target model. Training activations come
from FineWeb~\citep{penedo2024fineweb} \texttt{sample-10BT} documents with at least 128 tokens, truncated
to 1,024 tokens, with 10 token positions per document. Documents are split
disjointly into 25\% verbalizer supervised initialization, 25\% reconstructor
supervised initialization, and 50\% online training.

\begin{table}[h]
    \caption{Activation extraction sites. Revisions are Hugging Face commit
    prefixes; released NLAs are identified by their repository names. A
    released NLA at layer $K$ reads the output of decoder block $K$, i.e.,
    \texttt{hidden\_states[$K{+}1$]}.}
    \label{tab:activation-extraction}
    \begin{center}
    \begin{tabular}{@{}llll@{}}
        \toprule
        & Target model & Revision or release & Hidden state \\
        \midrule
        \multirow{3}{*}{Ours} & Qwen2.5-7B-Instruct & \texttt{a09a354} & \texttt{hidden\_states[21]} \\
        & Gemma-3-12B-it & \texttt{96b6f1e} & \texttt{hidden\_states[33]} \\
        & Apertus-v1.5-8B & \texttt{a411d83} & \texttt{hidden\_states[22]} \\
        \midrule
        \multirow{4}{*}{Released} & Qwen2.5-7B-Instruct & \texttt{nla-qwen2.5-7b-L20} & \texttt{hidden\_states[21]} \\
        & Gemma-3-12B-it & \texttt{nla-gemma3-12b-L32} & \texttt{hidden\_states[33]} \\
        & Gemma-3-27B-it & \texttt{nla-gemma3-27b-L41} & \texttt{hidden\_states[42]} \\
        & Llama-3.3-70B-Instruct & \texttt{Llama-3.3-70B-NLA-L53} & \texttt{hidden\_states[54]} \\
        \bottomrule
    \end{tabular}
    \end{center}
\end{table}

\paragraph{Evaluation token positions.}
For classification, the activation is extracted at the last source-text
token, before any evaluation question. For suffix prediction, it is
extracted at the last prefix token, before the true continuation. For
response behavior, it is extracted at the first token of a freshly generated
response from the target model. These positions are the same for all target
models. Explanations are sampled at temperature 1 with at most 150 new tokens.

\subsection{Evaluation protocols}
All judges, including the hard-negative judge
(Appendix~\ref{sec:hard-negative-details}), are DeepSeek V4.1 Flash~\citep{deepseekai2026v41flash} with
thinking disabled and temperature zero. Prompt templates are described in
Appendix~\ref{sec:judge-prompts}.

\paragraph{Utility tasks, controls, and aggregation.}
We evaluate utility on source-property classification,
suffix prediction, and response-behavior prediction. In all cases, the utility judge
receives the task query and the explanation.

\begin{table}[h]
\caption{Overview of our various utility tasks}
\label{tab:main3-utility}
\begin{center}
\setlength{\tabcolsep}{2pt}
\begin{tabular}{@{}llrrl@{}}
\toprule
Task & Judge's task & Examples & Options & Metric \\
\midrule
Classification & Source properties (12 tasks) & 1,344 & 2--14 & Mean accuracy \\
Suffix selection & True 32-token continuation & 500 & 10 & Accuracy \\
Response behavior & Answer, refuse, or clarify & 248 & 3 & Macro-F1 \\
Hard negatives & Contrastive question on source pair & 1,316 (658 pairs) & 2 & Log-prob.\ ratio \\
\bottomrule
\end{tabular}
\end{center}
\end{table}

For target instance $i$, let $q_i$ denote the task query,
$\mathcal{C}_i$ the candidate answers, $y_i$ the reference answer,
$z_i$ the matched explanation, and $z_{j(i)}$ the hard-negative
explanation, generated for the paired source $j(i)$.
Where predictive probabilities over $\mathcal{C}_i$ are available, we define
\begin{equation}
\begin{aligned}
L_{\mathrm{matched},i}
&=-\log p_{\mathrm J}(y_i\mid q_i,\mathcal C_i,z_i),\\
L_{\mathrm{noexp},i}
&=-\log p_{\mathrm J}(y_i\mid q_i,\mathcal C_i,\varnothing),\\
L_{\mathrm{hard},i}
&=-\log p_{\mathrm J}(y_i\mid q_i,\mathcal C_i,z_{j(i)}).
\end{aligned}
\label{eq:utility-judge-nll}
\end{equation}
We then decompose the hard-negative separation as
\begin{equation}
\begin{aligned}
U_{\mathrm{gain},i}
&=L_{\mathrm{noexp},i}-L_{\mathrm{matched},i},\\
U_{\mathrm{penalty},i}
&=L_{\mathrm{hard},i}-L_{\mathrm{noexp},i},\\
U_{\mathrm{contrast},i}
&=L_{\mathrm{hard},i}-L_{\mathrm{matched},i}
 =U_{\mathrm{gain},i}+U_{\mathrm{penalty},i}.
\end{aligned}
\label{eq:utility-decomposition}
\end{equation}
Thus, $U_{\mathrm{gain}}$ measures the predictive benefit of the matched
explanation over no explanation, whereas $U_{\mathrm{penalty}}$ measures
the degradation caused by the hard-negative explanation. Their sum is the
hard-negative separation. We report these components separately because
a larger separation may reflect more useful matched explanations, more
misleading hard negatives, or both. Appendix~\ref{sec:hard-negative-details} describes how hard-negative pairs
are constructed and reports the decomposition for all runs.

The classification suite contains 112 source examples for each of 12 tasks,
covering topic, sentiment, subjectivity, entailment, entity and exact-name
presence, tense, grammatical number, language, and pronoun form, for 1,344
examples per checkpoint and method. We average accuracy over successful
generations and parsed responses within each task and then weight the
12 tasks equally; details and data sources will be available in our repository upon acceptance.

Suffix prediction uses 500 documents with ten fixed candidate continuations
per example. The activation is extracted at the end of the prefix before
the true 32-token continuation.

Response-behavior prediction uses 248 LMSYS-Chat-1M~\citep{zheng2024lmsyschat} conversations. The
judge sees only the explanation and makes a forced prediction of
answer, refusal, or clarification. Reference labels come from each target
model's actual response. We report the macro-F1 score.

\paragraph{Source-support judgments.}
The source-support judge extracts distinct assertions and assigns supported, contradicted, unsupported, or not
assessed verdicts. Reasonable entailments count as supported; a
contradiction requires incompatible source evidence. Future-text
predictions and hidden-computation claims are not assessed and are excluded
from both metrics. Token coverage counts the union of tokens
overlapping unsupported or contradicted claims.

\paragraph{Writing defects.}
The writing judge sees only the explanation. It identifies distinct defects
in clarity, coherence, and economy, assigning each defect one category
and an exact evidence span. An annotation is accepted only when the quoted span occurs exactly in the explanation.

\paragraph{Checkpoint comparison.}
\label{sec:baseline-comparison-details}
Classification, suffix, and confabulation changes use each series' own
supervised initialization as reference. The models are
evaluated every 100 updates through 1600.
Tables~\ref{tab:baseline-comparison} and~\ref{tab:flow-model-comparison}
report absolute checkpoint scores rather than changes from supervised
initialization. We select one checkpoint per run by the highest unweighted
mean of classification accuracy, suffix accuracy, and response-behavior
macro-F1 among updates 1300--1600 for a fair comparison. Released checkpoints are evaluated
unchanged, with the same judges and protocols as our runs.
Classification, assertion risk, and token coverage weight the 12
classification tasks equally.

\section{Individual Evaluation Panels}
\label{sec:individual-evaluation-panels}
\begin{figure}[htbp]
    \centering
    \includegraphics[width=\linewidth]{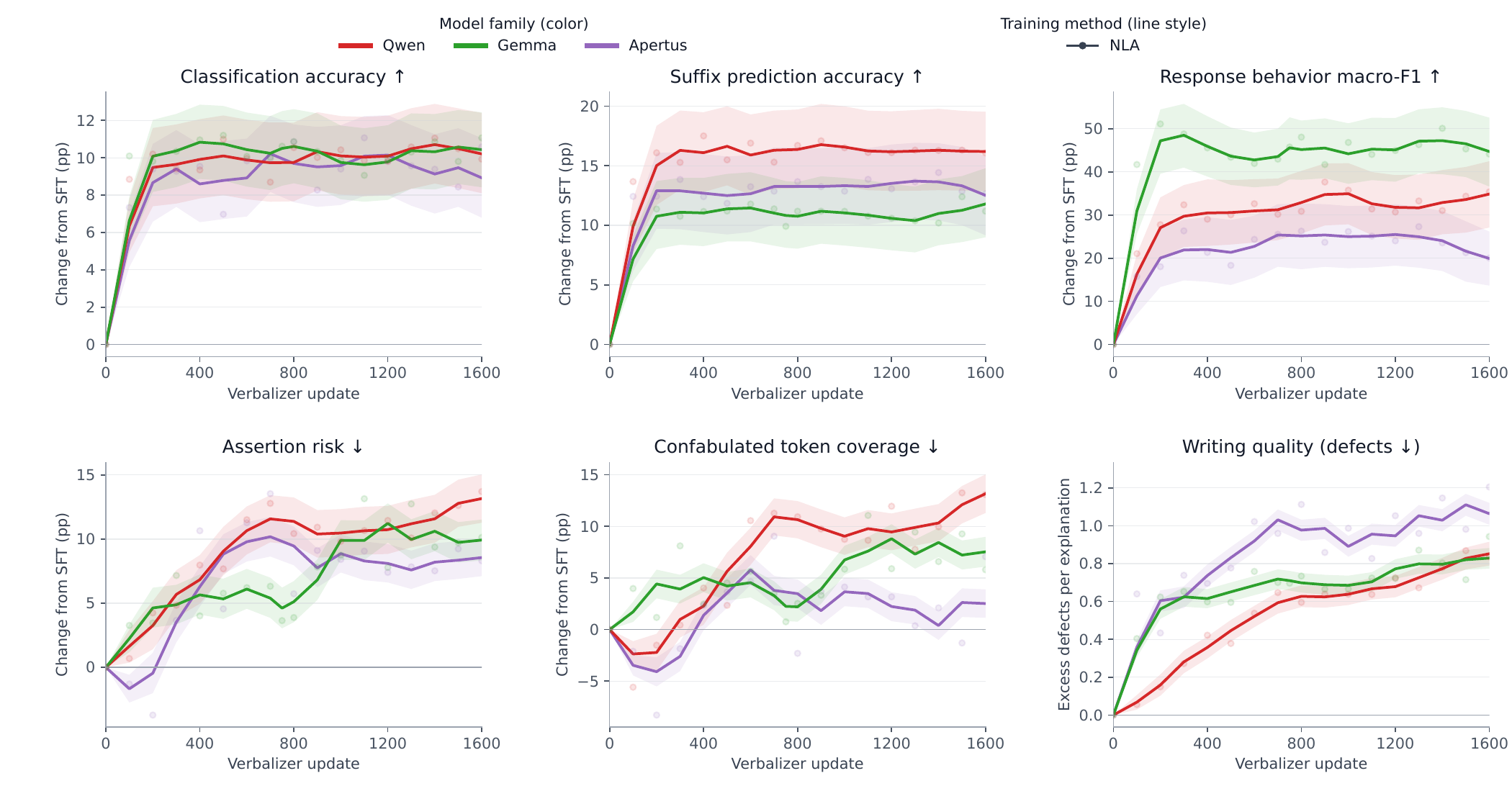}
    \caption{Changes in utility, confabulation, and writing defects from
    supervised initialization for the point-reconstruction NLA reruns. Shading shows
    pointwise 95\% bootstrap intervals.}
    \label{fig:individual-evaluation-panels}
\end{figure}

Figure~\ref{fig:individual-evaluation-panels} shows the individual utility,
confabulation, and writing trajectories of the point-reconstruction NLA
reruns summarized in Section~\ref{sec:training-trajectories}, including the
assertion-risk and writing-defect components of the confabulation index.

\section{Hard-Negative Separation}
\label{sec:hard-negative-details}
\paragraph{Pair construction.}
Hard-negative pairs are drawn from the classification sources. For each of
the 12 datasets, we retrieve lexical nearest
neighbors and write one question that the two sources answer differently
(factual, linguistic, or evaluative), with the two answers as candidates.
This gives 658 disjoint pairs covering 1,316 of the 1,344 classification
sources.

\paragraph{Judge and aggregation.}
The judge is DeepSeek V4.1 Flash with thinking disabled, as for the other
utility tasks. It answers with an option
letter, and we read the log-probability of the correct letter, normalized over
the two letters. Each question is scored with both answer orders to cancel
letter-position bias. Losses follow Eq.~\ref{eq:utility-judge-nll} and the
decomposition follows Eq.~\ref{eq:utility-decomposition}. We average within each dataset and then weight datasets equally.

\begin{figure}[htbp]
    \centering
    \includegraphics[width=\linewidth]{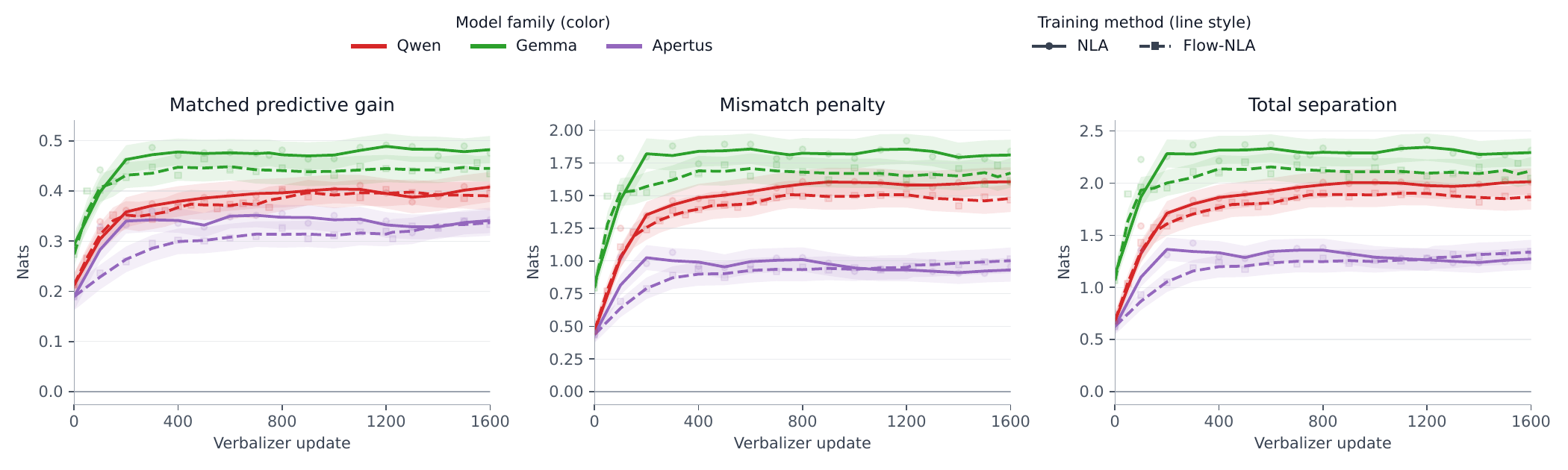}
    \caption{Matched predictive gain, mismatch penalty, and total
    hard-negative separation across the 12 classification datasets for
    point-reconstruction NLA (solid) and Flow-NLA (dashed). Curves are three-checkpoint means over raw
    markers; shading shows 95\% bootstrap intervals.}
    \label{fig:hard-negative-decomposition}
\end{figure}

\paragraph{Decomposition.}
Figure~\ref{fig:hard-negative-decomposition} separates the total separation
of Figure~\ref{fig:hard-negative-separation} into matched predictive gain
$U_{\mathrm{gain}}$ and mismatch penalty $U_{\mathrm{penalty}}$. Without an
explanation, the judge is at chance on every pair. Most of the separation,
and most of its rise during training, comes from the mismatch penalty: it
grows from 0.44--0.83 nats at supervised initialization to 0.91--1.86 nats
from update 800 onward, while the matched gain grows from 0.19--0.29 to
0.31--0.49 nats. Both components level off after about 200 updates. Late in
training, Flow-NLA matches NLA on both components for Apertus and has a
slightly lower mismatch penalty for Qwen and Gemma (by 0.10 and 0.16 nats).

\section{Training Details}
\label{sec:training-details}
Supervised initialization uses explanations written by Qwen3.6-35B-A3B for the
activation sites of Appendix~\ref{sec:activation-extraction}. Each run uses
8 NVIDIA GH200 GPUs. For reruns, we use the same hyperparameter settings as the released Anthropic code~\citep{frasertaliente2026nla}; for Apertus, we use the hyperparameters of Qwen2.5-7B as it is the closest in parameter size.

\section{Making the Distributional Reward Trainable}
\label{sec:reward-stability}
Using a co-trained likelihood bound as a policy-gradient reward raised
problems that a fixed reward model does not. This section records the
failure modes we encountered and how we addressed them. All diagnostics
reported here are for Qwen2.5-7B-Instruct.

\paragraph{The reconstructor tracks a moving target.}
The conditional distribution $p_\phi(h\mid z)$ changes with the
verbalizer, and the reconstructor measurably follows it. Between updates
0 and 200, the co-trained reconstructor's score improved by 38.8 units on
explanations from the update-200 verbalizer, but by only 5.7 units on
explanations from supervised initialization. A frozen reconstructor also confirms that the
verbalizer's gains are real: it scores update-200 explanations above
initial ones (0.466 versus 0.332). Monitoring with a fixed reconstructor
therefore measures transfer to old explanations, not joint progress.

\paragraph{Scoring with an averaged reconstructor.}
Unlike the released NLA recipe, which scores explanations with the current
reconstructor~\citep{frasertaliente2026nla}, we score them with an
exponential moving average (EMA) of the reconstructor's parameters with
decay 0.99. Rewards thus follow the moving conditional distribution
gradually rather than jumping with each reconstructor update. Such
parameter averaging is common in off-policy RL, where the targets that
critics are trained on are computed with an EMA of their
parameters~\citep{lillicrap2016continuous}.

\paragraph{Single noise levels give weak credit.}
An early reward used velocity error at a single uniformly sampled noise
level. At nearly clean inputs, the noisy activation reveals the target and
alternative explanations receive almost the same score. The gap between
mismatched and matched explanations was 0.003 at $t=0.05$ versus 0.625 at
$t=0.95$. At pure noise the reward reduces to point reconstruction
(Appendix~\ref{sec:statistical-foundations}). In the retained reward, the
variance that ranks candidate explanations comes mostly from intermediate
log-SNR: 2.3\% from $\lambda\le-4$, 34.9\% from $(-4,0]$, 60.9\% from
$(0,4]$, and 1.9\% from $\lambda>4$. Integrating across the band with
likelihood weighting captures this signal without choosing a single noise
level.

\paragraph{Monitoring distributional fit.}
\label{sec:distributional-fit}
Reconstruction FVE measures only how well the reconstructor recovers the
mean, so we also check whether the spread of its samples matches the
spread of the activations. On a fixed panel of held-out
activation--explanation pairs, we draw 16 samples per explanation by
running the reconstructor in reverse from noise. We compare the covariance
of these samples around their mean with the covariance of the activations
around the reconstructor's mean prediction, separately inside and outside
the subspace spanned by the 64 leading principal components of real
activations. The PCA-64 covariance overlap is one minus the
normalized squared Bures distance between the two covariances within
this subspace; it equals one when they are identical. The off-PC
variance ratio divides the samples' variance outside the subspace by
that of the activations, corrected for the finite number of samples; it
equals one when the model is calibrated there, and values below one
indicate samples that are too concentrated. For the reconstructor used by
the likelihood-bound reward, the off-PC variance ratio moves from 1.26 to
1.00 over reconstructor training steps 50--1{,}600, while the PCA-64
overlap rises from 0.77 to 0.89.

\paragraph{An energy-score reward.}
Before adopting the likelihood bound, we rewarded the verbalizer with the negative
energy score~\citep{gneiting2007proper} of $K=16$ reverse samples
$\widehat x^{(1)},\dots,\widehat x^{(K)}$ per explanation,
\begin{equation}
 \mathrm{ES}(x)
 =\frac1K\sum_{k}\|\widehat x^{(k)}-x\|_2
 -\frac{1}{2K(K-1)}\sum_{k\neq l}\|\widehat x^{(k)}-\widehat x^{(l)}\|_2,
\end{equation}
a proper scoring rule minimized in expectation by the true conditional
distribution, with common noise across candidates. Unlike the likelihood
bound, which only requires noising the observed activation forward, this
reward depends on samples generated by running the reconstructor in
reverse, so its value is only as good as the spread of those samples.
The reconstructor trained for this reward lost spread outside the leading
subspace: its off-PC variance ratio fell from 0.96 to 0.78 over
reconstructor training steps, while its PCA-64 overlap
stayed near 0.66 and FVE rose from 31.8\% to 49.6\%. Better mean
reconstruction thus came with increasingly concentrated samples during energy reward training.

\section{Judge Prompts}
\label{sec:judge-prompts}
The exact prompt templates will be available in the \texttt{prompts/} directory
of our evaluation repository upon acceptance. The utility judge receives the task
query, its candidate answers, and the explanation, and returns a single
answer; classification, suffix selection, and response behavior each have
their own template. The source-support judge receives the source text and
the explanation, splits the explanation into atomic claims, and returns a
verdict for each claim with quoted source evidence for supported and
contradicted claims. The writing judge receives only the explanation and
returns each defect with its category and quoted span. All judges return
structured output, which is validated before scoring.

\section{Why Denoising Identifies the Conditional Distribution}
\label{sec:statistical-foundations}
This section works through the example of
Figure~\ref{fig:equal-mean-reconstruction} and then shows that denoising at
an interior noise level identifies the explanation-conditioned distribution.

\paragraph{Illustrative example.}
In Figure~\ref{fig:equal-mean-reconstruction}, the activation is uniform over $(\pm1,0)$ and $(0,\pm1)$, and the explanation $z_{\mathrm{hor}}$ identifies the horizontal pair.
Both the unconditional and the $z_{\mathrm{hor}}$-conditional means are zero, so the optimal unit-norm point loss is $2$ with or without the explanation.
For this illustration, $x=h$ and $X_t=a_tX+b_t\epsilon$, with $\epsilon\sim\mathcal N(0,I)$ and $a_t^2+b_t^2=1$.
Panels (b)--(d) fix $b_t/a_t=0.6$ and use the observation $x_t=a_t(0.55,0.5)$ of the true activation $(1,0)$; shading shows the Gaussian-corrupted conditional densities.
Writing $m_z(x_t,t)=\mathbb E[X\mid X_t=x_t,Z=z]$, the field arrows in panels (c)--(d) depict
\begin{equation}
 d_z(x_t,t)=a_tm_z(x_t,t)-x_t
 =-b_t\widehat\epsilon^*(x_t,t,z)
 =b_t^2\nabla_{x_t}\log p_t(x_t\mid z).
\end{equation}
Thus, the displacement fields encode gradients of the noisy conditional log-density.
Flow-NLA scores noise predictions across noise levels using the likelihood bound in Appendix~\ref{sec:likelihood-bound}, allowing its reward to distinguish these equal-mean distributions.
With equally likely groups, group information leaves optimal point loss at $2$ but improves the optimal expected integrated denoising reward by approximately $\log 2$ nats over log-SNR $[-8,8]$, computed using exact posterior predictors and Gaussian quadrature (Appendix~\ref{sec:information-theory}).

\paragraph{Distribution identification.}
Fix the verbalizer, so that the joint activation--explanation distribution
is fixed. At an interior noise level with $a_t,b_t>0$, squared-error
training yields the population-optimal field
$f^*(x_t,t,z)=\mathbb E[v_t\mid x_t,z]$. Its implied noise prediction is
$\widehat\epsilon^*(x_t,t,z)=\mathbb E[\epsilon\mid x_t,z]$. Gaussian
corruption relates this prediction to the score of the noisy density,
$\nabla_{x_t}\log p_t(x_t\mid z)=-\widehat\epsilon^*(x_t,t,z)/b_t$,
where $\widehat\epsilon^*(x_t,t,z)=b_tx_t+a_tf^*(x_t,t,z)$.
 Its score determines it up to a normalization constant, fixed by requiring it to integrate to
one. Recovering the noisy distribution
therefore identifies the original conditional activation distribution at optimality.
 At the pure-noise endpoint $a_t=0$, the velocity target
is $-X$ and the optimum depends only on the conditional mean. Interior
noise levels are what allow equal-mean distributions to be distinguished.

\section{The Conditional Likelihood Bound}
\label{sec:likelihood-bound}
The reward in Eq.~\ref{eq:distributional-reward} is the negative interior
loss of a conditional diffusion variational bound
\citep{ho2020denoising,kingma2021variational,kingma2023understanding}.
We define the diffusion model on the chosen finite noise band, with an
explanation-independent Gaussian prior at the low-SNR boundary and an
explanation-independent observation decoder at the high-SNR boundary.
The explanation conditions the reverse transitions between these endpoints.
This specification ensures that the omitted endpoint terms do not vary
between candidate explanations of the same activation.

Because activations lie on a sphere, we take densities with respect to
the image $\nu$ of sphere surface measure under the fixed transform
$x=T(h)=(\sqrt d\,h-\mu)/s$. Following \citet{kingma2021variational}, the
decoder is $r(x\mid x_{\mathrm{hi}})\propto q(x_{\mathrm{hi}}\mid x)$,
normalized over $\nu$, where $q$ is the forward Gaussian corruption at the
high-SNR boundary.

Applying the standard variational bound separates the endpoint terms
from KL divergences between forward posteriors and learned reverse
transitions. For Gaussian reverse transitions with the forward posterior
variance, the interior terms are weighted squared noise-prediction errors.
In the continuous log-SNR limit, under the regularity and integrability
conditions for that limit, their sum becomes the negative of
Eq.~\ref{eq:distributional-reward}. Consequently,
$\log p_\theta(h\mid z)\ge R_\theta(h,z)-c(h)$,
where $c(h)$ contains the endpoint terms and is independent of $z$.
The bound applies to a well-defined reverse process without requiring
optimal denoising.

The velocity and noise parameterizations obey
$\epsilon-\widehat\epsilon_\theta=a_t(v_t-f_\theta)$, so the reward weights
velocity error by $a_t^2$ before integrating over log-SNR. This differs
from the reconstructor's coordinate-averaged training loss. Sampled noise
and numerical integration approximate the bound-derived reward; computing
candidate scores does not require evaluating the fixed observation decoder.

\section{Information Gain Across the Noise Band}
\label{sec:information-theory}
Fix the verbalizer and assume optimal denoisers with and without the
explanation. Let $\mathcal E_Z(\lambda)$ and $\mathcal E_0(\lambda)$ denote
their expected squared noise-prediction errors, and write
$Y_\lambda=e^{\lambda/2}X+\epsilon$, where $\epsilon$ is independent
standard Gaussian noise. Under finite second moments, the I-MMSE relation
\citep{guo2005mutual} gives
\begin{equation}
\begin{split}
 \frac12\int_{\lambda_{\min}}^{\lambda_{\max}}
 [\mathcal E_0(\lambda)-\mathcal E_Z(\lambda)]\,\mathrm d\lambda
 = I(Z;Y_{\lambda_{\max}})-I(Z;Y_{\lambda_{\min}}).
\end{split}
\end{equation}
The explanation's denoising benefit therefore measures information gained
across the selected noise band. An added detail can improve this quantity
by distinguishing modes or other distributional structure even when the
conditional mean remains unchanged. The equal-mean example in the main
text illustrates such a distinction. The finite band omits slight information already visible at its low-SNR boundary
and information still unresolved at its high-SNR boundary. The unconditional error contributes the same
term to every candidate score for a fixed activation and can be omitted
without changing candidate comparisons.

\end{document}